\documentclass[runningheads]{llncs}
\usepackage[T1]{fontenc}
\usepackage{graphicx}
\usepackage{hyperref}       
\usepackage{url}
\usepackage{caption}        
\usepackage{subcaption}
\usepackage{booktabs}       
\usepackage{makecell}
\usepackage{multirow}

\usepackage{color}

\usepackage{amsmath}
\usepackage{amssymb}
\usepackage{mathtools}

\begin{document}
\title{All Models Are Miscalibrated, But Some Less So: Comparing Calibration with Conditional Mean Operators}
\titlerunning{Comparing Calibration with Conditional Mean Operators}
%
\author{Peter Moskvichev \and
Dino Sejdinovic}
\authorrunning{P. Moskvichev and D. Sejdinovic}
\institute{University of Adelaide, Australia \\
\email{\{peter.moskvichev,dino.sejdinovic\}@adelaide.edu.au}}

\maketitle              

\begin{abstract}
When working in a high-risk setting, having well calibrated probabilistic predictive models is a crucial requirement. However, estimators for calibration error are not always able to correctly distinguish which model is better calibrated. We propose the \emph{conditional kernel calibration error} (CKCE) which is based on the Hilbert-Schmidt norm of the difference between conditional mean operators. By considering calibration error as the distance between conditional distributions, which we represent by their embeddings in reproducing kernel Hilbert space, the CKCE is less sensitive to the marginal distribution of predictive model outputs. This makes it more effective for relative comparisons than previously proposed calibration metrics. Our experiments, using both synthetic and real data, show that CKCE provides a more reliable measure of a model's calibration error and is more robust against distribution shift. 

\end{abstract}

\section{Introduction} \label{sec:intro}

Probabilistic models quantify uncertainty by predicting a probability distribution over possible labels, rather than simply giving a point estimate. This allows models to express their confidence in a prediction. However, to ensure that the probability outputs can be trusted by users, such models must satisfy certain calibration (or reliability) properties. Roughly speaking, calibration requires that predicted probabilities match observed frequencies of labels. A classical example is in weather forecasting. A forecaster that assigns 70\% chance of rain is well calibrated if, on average, it actually rains on 70\% of days with such predictions. Although calibration was originally studied in a meteorological context \cite{Brier_1950,DeGroot_1983}, it has proven to be useful in a wide range of applications. 


Work by \cite{Guo_2017} cast a spotlight on the miscalibration of deep neural networks, which often give overconfident predictions. More recently, \cite{phan2025hle} showed similarly poor calibration of state-of-the-art large language models. This gives reason to be concerned about the use of advanced AI models in high-risk settings. However, properly measuring calibration remains a challenge, with most authors and practitioners resorting to proxy metrics that do not fully capture the notion of \emph{strong calibration} explained in Section \ref{sec:background}. 

\begin{figure}[t]
    \centering
    \begin{minipage}{0.6\textwidth}
        \centering
        \includegraphics[width=\linewidth]{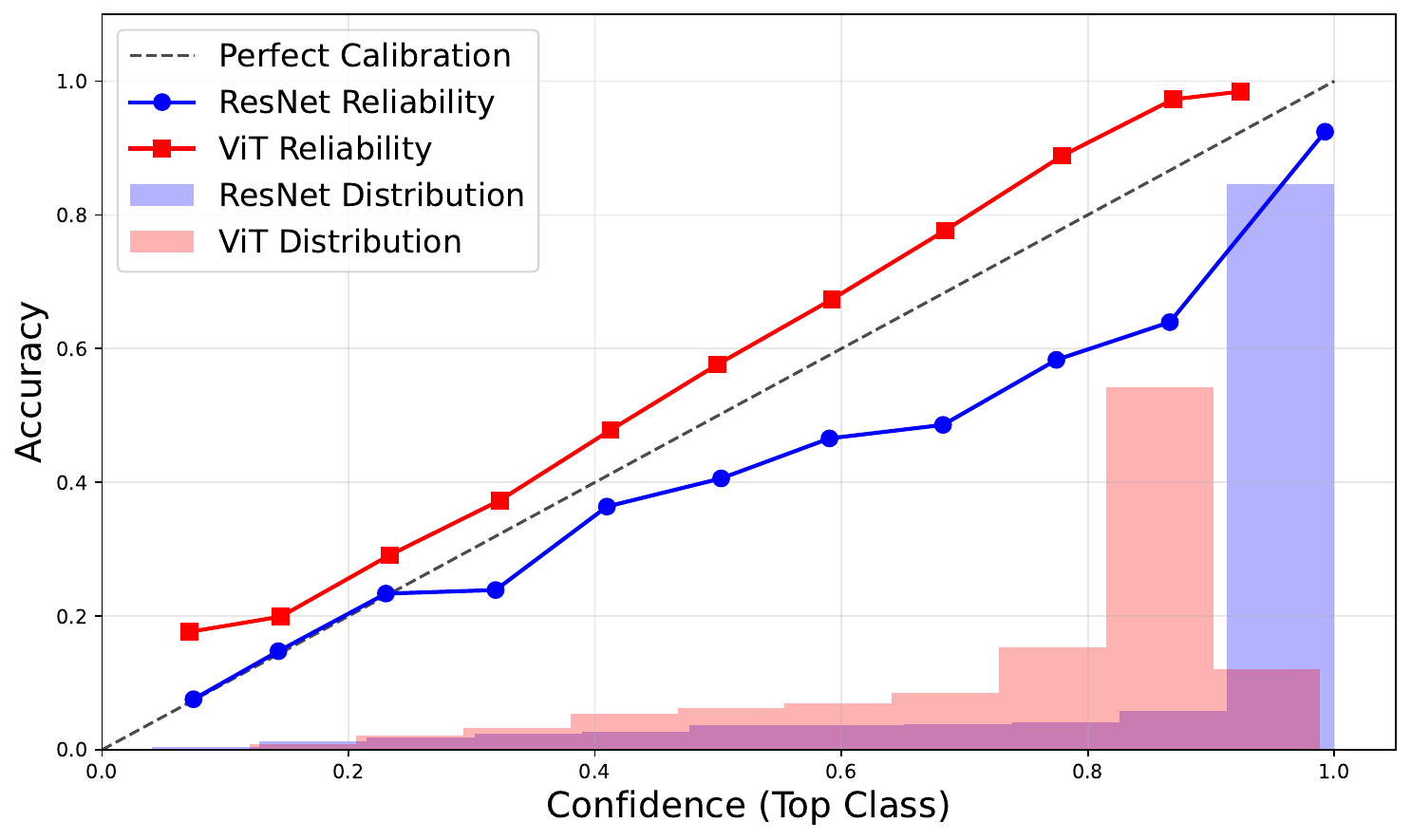}
    \end{minipage}
    \begin{minipage}{0.39\textwidth}
        \centering
        \begin{tabular}{lcc}
            \multicolumn{3}{c}{Calibration Error} \\
            \toprule
             & ECE & CKCE (ours) \\
            \midrule
            ResNet & 0.316 & 0.061 \\
            ViT & 0.389 & 0.044 \\
            \bottomrule
        \end{tabular}
    \end{minipage}
    \caption{Reliability diagram and distribution of prediction confidence for two models trained on ImageNet. A lower CKCE value indicates a closer match between model confidence and accuracy, whereas ECE is heavily affected by the marginal distribution of predictions.}
    \label{fig:reliability-diag}
\end{figure}

Reliability diagrams \cite{DeGroot_1983}, such as in Figure \ref{fig:reliability-diag}, offer a method of inspecting calibration by comparing the model's confidence in the predicted label and its accuracy, which should be equal for well calibrated forecasters. While reliability diagrams provide a useful visual tool, they fail to quantify calibration error, and are difficult to adapt to strong calibration. A popular calibration metric inspired by reliability diagrams is the expected calibration error (ECE) \cite{naeini_2015} which measures the expected difference between the predictive confidence and observed frequencies of labels. But this quantity is heavily influenced by the marginal distribution of model predictions, making direct model comparison difficult. Furthermore, estimating ECE can be problematic as it is highly sensitive to the chosen binning scheme which leads to biased and inconsistent estimators \cite{Nixon_2019,vaicenavicius_2019,Widmann_2019}. 

An alternative approach to calibration is via finding discrepancies between probability distributions, which can be represented by their kernel mean embeddings in a reproducing kernel Hilbert space (RKHS). This idea was used by \cite{Widmann_2021} to measure calibration via the difference in embeddings of joint distributions. While this is an effective tool for the purposes of a calibration hypothesis test, to which much work has been devoted \cite{Widmann_2021,Lee_2023,Glaser_2023,Chatterjee_2024}, the proposed estimator is once again sensitive to the marginal distribution of predictive confidence. Therefore, just like the classical ECE, the calibration error estimator proposed by \cite{Widmann_2019} can be ill-suited for comparing different candidate models, particularly in situations with distribution shift. 

In this paper, we introduce the \emph{conditional kernel calibration error} (CKCE). It evaluates calibration by determining the discrepancy between conditional distributions which we represent by their RKHS conditional mean operator. This can be viewed as complementary to the work of \cite{Widmann_2021}. It might be challenging to use CKCE as a test statistic for a calibration test because of the difficulty in estimating the null distribution, but we are interested in a different problem: reliable relative calibration comparisons between models. We demonstrate through a range of experiments on both synthetic and real data that the CKCE is a more robust metric for this task.

\section{Background} \label{sec:background}

\subsection{Probabilistic Models and Calibration} \label{subsec:calib}

Consider a classification problem in which we aim to predict a label $y \in \mathcal{Y} = \{1, \dots, m\}$ based on a feature $x \in \mathcal{X}$. Rather than making a point prediction, we fit a probabilistic classifier $f: \mathcal{X} \to \Delta^{m-1}$ which estimates the conditional distribution $P(Y|X=x) \in \Delta^{m-1}$, where $\Delta^{m-1}$ represents the $(m-1)$-dimensional probability simplex.

A desirable property for the classifier is that the confidence in the predicted class matches the proportion of times the prediction is correct for that confidence. This is known as \emph{confidence calibration} and can be expressed as
\begin{equation} \label{eq:conf_calib}
    P(Y = \arg \max f(X) | \max f(X) = c) = c
\end{equation}
for all $c \in [0,1]$ \cite{Guo_2017,vaicenavicius_2019}. However, (\ref{eq:conf_calib}) does not consider calibration of \emph{all classes}, which may be crucial if the entire predicted probability vector is used for decision making or as an intermediate output for downstream tasks, particularly when certain classes carry varying levels of risk for those tasks. Therefore, our focus is on the notion of so called \emph{strong calibration} \cite{Widmann_2019}.
\begin{definition} \label{def:calibration}
    A probabilistic classifier $f: \mathcal{X} \to \Delta^{m-1}$ is \emph{strongly calibrated} if 
    \begin{equation} \label{eq:calibration}
        P(Y=i | f(X) = q) = q^{(i)} \qquad \text{for } i=1,\dots m
    \end{equation}
    for all $q = (q^{(1)}, \dots, q^{(m)}) \in f(\mathcal{X}) \subseteq \Delta^{m-1}$.
\end{definition}
We note that calibration is a property of a model that is essentially orthogonal to that of accuracy. Indeed, a predictive model which for any input simply returns the marginal distribution of class labels, $f(x)=P(Y)$, is calibrated according to Definition \ref{def:calibration}, but has accuracy at a chance level. Conversely, a classifier may attain the optimal (Bayes) risk with respect to the 0/1 loss (the highest possible accuracy) but have arbitrarily bad calibration. An example would be a classifier that assigns all probability mass to label $\hat Y=\arg\max_{i=1,\ldots,m}P(Y=i|X)$.

If we let $Q_X = f(X)$, then the discrepancy between the probability distributions $P(Y|Q_X)$ and $Q_X$ (corresponding to the left and right hand sides of \eqref{eq:calibration} in Definition \ref{def:calibration}) can be understood as the \emph{calibration error} of the probabilistic model. However, while $Q_X$ is readily available from the model output, $P(Y|Q_X)$ can be highly complex, making it difficult to estimate. This motivates the use of nonparametric methods which allow to represent probability distributions and distances between them using only samples from the distributions.

\subsection{Kernel Embeddings of Distributions} \label{sec:kernel_emb}

We provide a short review of some preliminaries on kernel embeddings of probability distributions in a reproducing kernel Hilbert space (RKHS) \cite{muandetKernelMeanEmbedding2017a}. Consider a random variable $X$ with domain $\mathcal{X}$ and distribution $P(X)$. Let $k:\mathcal{X} \times \mathcal{X} \to \mathbb{R}$ be any positive definite function with associated RKHS $\mathcal{H}_k$ and canonical feature map $\phi(x) = k(x, \cdot)$. 

Assuming $k$ is bounded, the \emph{kernel mean embedding} of $P(X)$ is given by $\mu_{X} = \mathbb{E}_X k(X, \cdot) \in \mathcal{H}_k$ and has the property $\langle \mu_{X}, f \rangle_{\mathcal{H}_k} = \mathbb{E}_X f(X)$ for all $f \in \mathcal{H}_k$. Given another random variable $Y$ with embedding $\mu_{Y}$, we can compute the distance between $P(X)$ and $P(Y)$ using the \emph{maximum mean discrepancy} (MMD)
\begin{equation}
\begin{aligned}
    \text{MMD}_k^2(P(X), P(Y)) &= \| \mu_{X} - \mu_{Y} \|_{\mathcal{H}_k}^2 \\
    &= \mathbb{E}_{X, X'} k(X, X') + \mathbb{E}_{Y, Y'} k(Y, Y') -2\mathbb{E}_{X, Y} k(X, Y)
\end{aligned}
\end{equation}
The MMD can be easily estimated using samples from $P(X)$ and $P(Y)$ \cite{Gretton_2012}. For so called \emph{characteristic} kernels, the mean embeddings are injective and the MMD equals zero if and only if the distributions $P(X)$ and $P(Y)$ are equal \cite{Fukumizu_2007}. Many commonly used kernels such as the Gaussian and Laplace kernels are characteristic. A closely related notion is a \emph{universal} kernel, which when defined on a compact metric space $\mathcal{X}$, has an associated RKHS $\mathcal{H}$ that is dense in the space of continuous functions on $\mathcal{X}$ \cite{sriperumbudurUniversalityCharacteristicKernels2011}.

We can similarly define the kernel embedding of a joint distribution $P(X, Y)$ by choosing a kernel on the product domain $\mathcal{X} \times \mathcal{Y}$. Given kernels $k$ and $\ell$ on $\mathcal{X}$ and $\mathcal{Y}$ respectively, with associated RKHS $\mathcal{H}_k$ and $\mathcal{H}_\ell$, and defining the tensor product kernel $(k \otimes \ell)((x,y),(x',y')) = k(x,x')\ell(y,y')$, a special case for the joint mean embedding is $\mu_{X Y} = \mathbb{E}_{XY} [\phi(X) \otimes \psi(Y)] \in \mathcal{H}_{k\otimes \ell}$ where $\phi(x) = k(x, \cdot)$, $\psi(y) = \ell(y, \cdot)$ are the canonical feature maps \cite{Fukumizu_2004}. By isometry between $\mathcal{H}_{k\otimes \ell}$ and $\mathcal{H}_{k}\otimes \mathcal{H}_\ell$, the joint mean embedding $\mu_{XY}$ can be identified with the uncentered cross-covariance operator $\mathcal{C}_{XY}: \mathcal{H}_\ell \to \mathcal{H}_k$, which has the property $\mathbb E[f(X)g(Y)] = \langle f , \mathcal{C}_{XY}g \rangle_{\mathcal{H}_k}$.

\subsection{Conditional Mean Operators}

Conditional distributions $P(Y|X=x)$ can likewise be embedded in an RKHS by the conditional mean embedding $\mu_{Y|X=x} = \mathbb{E}_{Y|X = x} [\psi(Y) | X=x] \in \mathcal{H}_\ell$ \cite{Song_2009}. Note that the conditional mean embedding defines a family of points in the RKHS indexed by the values $x$. This motivates us to consider the \emph{conditional mean operator} (CMO) $\mathcal{C}_{Y|X}:\mathcal H_{k}\to\mathcal H_{\ell}$ which satisfies $\mu_{Y|X=x} = \mathcal{C}_{Y|X}k(x, \cdot)$. Assuming that for all $g \in \mathcal H_{\ell}$, we have $\mathbb{E}_Y[g(Y) | X = \cdot] \in \mathcal H_k$, the CMO is a well defined Hilbert-Schmidt operator.
Much research has been devoted to put conditional mean embeddings and operators on a rigorous footing, including the measure theoretic view taken by \cite{Park_2020}. Given a dataset $\{(x_i, y_i)\}_{i=1}^n$ sampled from $P(X,Y)$, an empirical estimator of the CMO is 
\begin{equation} \label{eq:cmo_est}
    \widehat{\mathcal{C}}_{Y|X} = \Psi_Y (K_{XX} + \lambda n I_n)^{-1} \Phi_X^*
\end{equation}
where $^*$ indicates the adjoint of an operator, $\Psi_Y = [\psi (y_1), \dots, \psi (y_n)]$, $\Phi_X = [\phi(x_1), \dots, \phi(x_n)]$, $K_{XX} = \Phi_X^* \Phi_X$ is the Gram matrix with entries $[K_{XX}]_{ij} = k(x_i, x_j)$, $I_n$ is the $n\times n$ identity matrix and $\lambda$ is a regularisation parameter \cite{Song_2009}. Assuming that $\mathcal{C}_{YX} \mathcal{C}_{XX}^{-3/2}$ is Hilbert-Schmidt, then $\| \widehat{\mathcal{C}}_{Y|X} - \mathcal{C}_{Y|X} \|_{HS} = O_p(\lambda^{1/2}+\lambda^{-3/2}n^{-1/2})$. Therefore, if the regularisation term satisfies $\lambda \to 0$ and $n \lambda^3 \to \infty$, then $\widehat{\mathcal{C}}_{Y|X}$ converges in probability to the true CMO \cite[Theorem 1]{Song_2010}. A detailed refined analysis of convergence of conditional mean embeddings is studied in \cite{Li_2022}. Additionally, \cite[Supplementary Material, Section 8]{singh_2023} provides a minimax optimal rate for the convergence of CMO in Hilbert-Schmidt norm, which can depend on kernel $k$ and the smoothness of underlying conditional distributions. These results can provide a recipe for selecting the schedule of the regularisation parameter $\lambda$. An alternative method is to view the CMO as the solution to a regression problem, allowing $\lambda$ to be selected via cross-validation \cite{Grunewalder_2012}. For simplicity, we choose the regularisation parameter to have schedule $\lambda_n = n^{-1/4}$.

Consider two random variables $Y$ and $Z$ with common domain $\mathcal{Y}$ and a conditioning random variable $X$. Assuming both CMOs are well defined, if $\mathcal{C}_{Y | X} = \mathcal{C}_{Z | X}$, then the conditional distributions $P(Y|X)$ and $P(Z|X)$ are equal in the sense that $P(Y|X=x) = P(Z|X=x)$ for $P(X)$-almost all $x \in \mathcal{X}$ \cite[Theorem 3]{Ren_2016}. Hence, the distance between conditional probability distributions is measured by the \emph{conditional maximum mean discrepancy} (CMMD) \cite{Ren_2016} which is defined as
\begin{equation}
    \text{CMMD}_{k, \ell}^2 (P(Y|X), P(Z|X)) = \| \mathcal{C}_{Y | X} - \mathcal{C}_{Z | X}\|_{HS}^2,
\end{equation}
where $k$ and $\ell$ are kernels on $\mathcal{X}$ and $\mathcal{Y}$ respectively, and $\|\cdot\|_{HS}$ represents the Hilbert-Schmidt norm of an operator. In practice, empirical CMO estimators from equation (\ref{eq:cmo_est}) can be used to estimate the CMMD.

\section{Conditional Kernel Calibration Error} \label{sec:ckce}

Recall our notion of calibration error described in Section \ref{subsec:calib} where we compare the difference between probability distributions $P(Y|Q_X)$ and $Q_X$. By defining a random variable $Z_X \sim Q_X$ obtained from the predictive model, we note that $Q_X = P(Z_X | Q_X)$ and hence, a probabilistic predictive model is calibrated if and only if $P(Y|Q_X) = P(Z_X|Q_X)$ almost surely. Since we are now comparing differences between conditional probability distributions, we apply the CMMD and define the \emph{conditional kernel calibration error} (CKCE) with respect to kernels $k: \Delta^{m-1} \times \Delta^{m-1} \to \mathbb{R}$ and $\ell: \mathcal{Y} \times \mathcal{Y} \to \mathbb{R}$ as
\begin{equation}
    \text{CKCE}_{k, \ell} = \| \mathcal{C}_{Y | Q_X} - \mathcal{C}_{Z_X | Q_X} \|_{HS}^2.
\end{equation}
A CKCE of 0 implies the conditional distributions are equal almost everywhere, and thus strong calibration holds. Since the CKCE is not impacted by the marginal distribution of $Q_X$, it provides a measure of calibration error for relative comparisons of candidate models.

\subsection{Empirical Evaluation}

The CKCE can be computed by using empirical estimates of the CMOs as in equation (\ref{eq:cmo_est}). Given a set of $n$ calibration samples $\{(x_i,y_i) \}_{i=1}^n$ from the joint distribution $P(X, Y)$, we compute prediction $q_i = f(x_i)$ for each feature using the classifier. Then $\{(y_i, q_i) \}_{i=1}^n$ are samples from $P(Y, Q_X)$, and the CMO $\mathcal{C}_{Y|Q_X}$ is estimated by $\widehat{\mathcal C}_{Y | Q_X} = \Psi_Y (K_{QQ} + \lambda n I_n)^{-1} \Phi_Q^*$.

The CMO $\mathcal{C}_{Z_X|Q_X}$ can be estimated in a similar way. However, rather than sampling $Z_{x_i}$ from model outputs $q_i$ to construct the feature matrix $\Psi_{Z_X}$, we can replace each of the feature embeddings $\psi(Z_{x_i})$ with their expected value, that is, the kernel mean embedding $\mu_{Z_{x_i}}$. This reduces variance of the CMO estimator\footnote{A similar strategy was used in \cite{ChaBouSej2021} and termed \textit{Conditional Mean Shrinkage Operator}.}. Hence, $\widehat{\mathcal C}_{Z_X | Q_X} = M_Z (K_{QQ} + \lambda n I_n)^{-1} \Phi_Q^*$ where $M_Z = [\mu_{Z_{x_1}}, \dots, \mu_{Z_{x_n}}]$ is a matrix of mean embeddings of the predictive distributions. Using the CMO estimators, the empirical evaluation of the CKCE is given by
\begin{equation}
\begin{aligned}
    \widehat{\text{CKCE}}_{k, \ell} &= \| \widehat{\mathcal C}_{Y | Q_X} - \widehat{\mathcal C}_{Z_X | Q_X} \|_{HS}^2 \\
     &= Tr\Big(  W[M_Z^* M_Z + L_{YY} - M_Z^* \Psi_Y - \Psi_Y^* M_Z] W K_{QQ} \Big)
\end{aligned}
\end{equation}
where $W = (K_{QQ} + \lambda n I_n)^{-1}$, $L_{YY} = \Psi_Y^* \Psi_Y$ is the Gram matrix with entries $[L_{YY}]_{ij} = \ell(y_i, y_j)$, and we have applied the fact that the squared Hilbert-Schmidt norm of an operator can be expressed as $\|\mathcal{C}\|_{HS}^2 = Tr(\mathcal{C}^* \mathcal{C})$.

Although estimating the CKCE involves finding the inverse of a potentially large matrix, which is computationally expensive, this can be overcome by a suite of well established large-scale kernel approximations such as random Fourier features \cite{rahimiRandomFeaturesLargescale2007}, which we explore in Section \ref{sec:rff}. Alternatively, simple low-rank approximations of the Gram matrix $K_{QQ}$ such as incomplete Cholesky factorization \cite{Fine_2002} can be used for further computational efficiency.

\subsection{Choice of Kernel} \label{sec:choice_kernel}

Evaluating the CKCE requires choosing two kernels, one on the space of labels $\mathcal{Y}$ and another on the probability simplex $\Delta^{m-1}$. A reasonable choice for the former is the Kronecker kernel $\ell(y,y') = \mathbf{1}\{ y = y' \}$. In this case $\mathcal{H}_\ell \cong \mathbb{R}^m$ with feature maps $\psi(j) = e_j$ as the canonical basis. Then, for a conditional distribution $P(Y|Q_X)$ described by $\pi(q) = P(Y | Q_X =q)$, we have $\mu_{Y|Q_X=q} = \mathbb{E}[e_Y | Q_X=q] = \sum_{j=1}^m \pi_j(q) e_j = \pi(q)$, which is understood as a vector in $\mathbb{R}^m$. Thus, the conditional mean embedding is some mapping (potentially nonlinear) of the probability vector $q$. On the other hand, $P(Z_X | Q_X =q) = q$, and so the conditional mean embedding is simply $\mu_{Z_X|Q_X=q} = q$. 

Recall our assumption for well defined CMOs that for all $g \in \mathcal{H}_\ell$, we have $\mathbb{E}_Y[g(Y)|Q_X = \cdot]$, $\mathbb{E}_{Z_X}[g(Z_X)|Q_X = \cdot] \in \mathcal{H}_k$. Since by isometry $\mathcal{H}_\ell \cong \mathbb{R}^m$ each $g$ has a corresponding vector $v \in \mathbb{R}^m$, we get $\mathbb{E}_Y[g(Y)|Q_X = q] = v^\top \pi(q)$ and $\mathbb{E}_{Z_X}[g(Z_X)|Q_X = q] = v^\top q$. By choosing a universal kernel $k$ on $\Delta^{m-1}$, $\mathcal{H}_k$ is rich enough to approximate functions $v^\top \pi(q)$ arbitrarily well. At the same time, the RKHS should also contain linear maps to have $v^\top q \in \mathcal{H}_k$. Combing these two desiderata, we propose the kernel $k: \Delta^{m-1} \times \Delta^{m-1} \to \mathbb{R}$ with
\begin{equation} \label{eq:kernel}
    k(p,q) = p^\top q + \exp\left( -\frac{1}{2 \gamma^2} \| p - q \|^2 \right)
\end{equation}
where $\gamma \in \mathbb{R}$ is the bandwidth. Such a combination of the linear and Gaussian kernels ensures that the CMO assumptions are reasonably satisfied, and establishes robustness of the CKCE metric which we demonstrate experimentally in Section \ref{sec:robustness}.


\subsection{Random Fourier Feature Implementation} \label{sec:rff}

An additional benefit of kernel (\ref{eq:kernel}) is that it can be easily approximated by random Fourier features. This allows each CMO to be estimated in the primal by applying the Woodbury identity, which speeds up computation for large $n$ without sacrificing convergence properties \cite{liUnifiedAnalysisRandom2021}. Since adding kernels is equivalent to concatenating features, an approximation of the canonical feature map for kernel $k$ is 
\begin{equation*}
    \hat \phi(p) = \left[p^\top, \frac{1}{\sqrt{D}} \cos ( w_1^\top p), \frac{1}{\sqrt{D}} \sin ( w_1^\top p ), \dots, \frac{1}{\sqrt{D}} \cos (w_D^\top p), \frac{1}{\sqrt{D}} \sin (w_D^\top p) \right]^\top
\end{equation*}
where $p \in \Delta^{m-1}$ and $w_1, \dots, w_D$ are $D$ independent samples from the multivariate normal distribution $\mathcal{N}_m(0, \gamma^{-2} I)$. One can show that $\mathbb{E}[ \hat \phi(p)^\top \hat \phi(q)] = k(p, q)$ and $\text{Var}[ \hat \phi(p)^\top \hat\phi(q)] = O(D^{-1})$ \cite{rahimiRandomFeaturesLargescale2007}. In our experiments, we choose $D=100$. 

Given a calibration set $\{(y_i , q_i)\}_{i=1}^n$, where each $q_i$ is the probability vector predicted by our classifier for a feature $x_i$, the CMO estimators are
\begin{equation}
    \widehat{\mathcal C}_{Y | Q_X} = \Psi_Y \widehat\Phi_Q^* (\widehat\Phi_Q \widehat\Phi_Q^* + \lambda n I_{\tilde m})^{-1}, \qquad
    \widehat{\mathcal C}_{Z_X | Q_X} = M_Z \widehat\Phi_Q^* (\widehat\Phi_Q \widehat\Phi_Q^* + \lambda n I_{\tilde m})^{-1} 
\end{equation}
where $\widehat \Phi_Q = [\hat\phi(q_1) \dots \hat\phi(q_n)]$ and $I_{\tilde m}$ is the identity matrix with $\tilde m = m+2D$. The resulting CKCE estimator is $||\widehat{\mathcal C}_{Z_X | Q_X} - \widehat{\mathcal C}_{Y | Q_X}||_F^2$ where we have replaced the Hilbert-Schmidt norm with the Frobenius norm as we are now working with finite dimensional matrices rather than operators. As the main computational complexity comes from calculating the inverse, we have thus reduced the computational cost from $O(n^3)$ to $O(\tilde m^3)$ with $\tilde m \ll n$.

\section{Comparison with other Calibration Measures}

\subsection{Proper Scoring Rules}

Historically, proper scoring rules such as the Brier score have been used to determine calibration \cite{Brier_1950}. Other common proper scoring rules are the negative log likelihood and cross entropy, which are both frequently used as loss functions when training modern machine learning models. Since these loss functions are minimised by the true conditional distribution, they can give a well calibrated model with enough training data. However, complex models can achieve low loss as a result of high accuracy rather than good calibration, meaning these metrics are of limited use for measuring calibration error directly. Furthermore, \cite{Merkle_2013} showed that different scoring rules can rank the same models differently, hampering their use for relative comparison of calibration.

\subsection{Expected Calibration Error}

A widely used measure of calibration for probabilistic models is the \emph{expected calibration error} (ECE) \cite{naeini_2015}. Although in its usual formulation ECE measures confidence calibration of probabilistic classifiers, it can be generalised to strong calibration as given by Definition \ref{def:calibration}. Let $d$ be a metric on $\Delta^{m-1}$. Then the ECE of a probabilistic model with respect to $d$ is defined as
\begin{equation}
    \text{ECE}_d = \mathbb{E}_X d(P(Y|Q_X), Q_X).
\end{equation}
The main difficulty in computing the ECE comes from estimating $P(Y|Q_X)$. An approach suggested by \cite{vaicenavicius_2019} is to partition the probability simplex and use histogram regression, which we implemented in our experiments. The simplest choice is uniform binning, although this causes the number of bins to increase exponentially with the number of classes, meaning care must be taken to ensure enough samples per bin. Other binning schemes may also be used, but as demonstrated by \cite{vaicenavicius_2019}, the ECE estimator becomes highly sensitive to the chosen scheme further emphasising the need for an alternative calibration metric. A special case of ECE is the average distance between conditional mean embeddings, which is achieved by choosing $d$ to be the squared MMD. This approach was further explored by \cite{Chatterjee_2024}, but their method requires taking samples $Z_X$ from $Q_X$ which can be difficult to marginalise, adding extra variance to the estimator.

\subsection{Joint Kernel Calibration Error}

An alternative kernel based measure of calibration considers the discrepancy between joint distributions. As before, we define the random variable $Z_X \sim Q_X$. Then the joint distributions $P(Y, Q_X)$ and $P(Z_X, Q_X)$ are equal if and only if $P(Y | Q_X) = Q_X$ almost surely, that is, the probabilistic predictive model is calibrated \cite{Widmann_2021}. Let $\kappa: (\mathcal{Y} \times \Delta^{m-1}) \times (\mathcal{Y} \times \Delta^{m-1}) \to \mathbb{R}$ be a measurable kernel with associated RKHS $\mathcal{H}_{\kappa}$. Then, the \emph{joint kernel calibration error} (JKCE) with respect to kernel $\kappa$ is defined as
\begin{equation} \label{eq:jkce}
    \text{JKCE}_\kappa = \| \mu_{Y Q_X} - \mu_{Z_X Q_X} \|_{\mathcal{H_{\kappa}}}^2.
\end{equation}
The JKCE corresponds to the MMD between the joint distributions $P(Y, Q_X)$ and $P(Z_X, Q_X)$. As described in Section \ref{sec:kernel_emb}, for a characteristic kernel $\kappa$ the JKCE is 0 if and only if the model is calibrated. Given kernels $\kappa_{\mathcal{Y}}$ on the space of labels and $\kappa_{\Delta}$ on the probability simplex, a kernel on the product domain $\mathcal{Y} \times \Delta^{m-1}$ can be constructed by their tensor product. If $\kappa_{\mathcal{Y}}$ and $\kappa_{\Delta}$ are universal kernels on locally compact Polish spaces, then $\kappa = \kappa_{\mathcal{Y}} \otimes \kappa_{\mathcal{P}}$ is characteristic \cite[Theorem 5]{Szabo_2018}. Since the kernels $k$ and $\ell$ from Section \ref{sec:choice_kernel} are universal, setting $\kappa = \ell \otimes k$ produces a characteristic kernel. We will use this kernel to evaluate JKCE in our experiments in Section \ref{sec:experiments} for a fair comparison with CKCE. 


\section{Experimental Results} \label{sec:experiments}

In our experiments, we assess the ability of the CKCE to distinguish between models through a relative comparison of calibration error. We compare its performance with ECE and JKCE. In the following experiments, the kernel in equation (\ref{eq:kernel}) is used on the input variable for both the CKCE and JKCE, and the bandwidth parameter $\gamma$ is chosen via the median heuristic \cite{Gretton_2012}. The ECE is estimated by uniform binning of the simplex, and we take $d$ to be the $L^1$ norm as is common in the literature \cite{Guo_2017,vaicenavicius_2019}.

\begin{table*}[t]
    \centering
    \caption{Performance metrics and calibration error of neural network architectures and the marginal probability model on the ImageNet dataset.}
    \begin{tabular}{cc|ccccc}
    \toprule
Data & Model & Accuracy & Cross Entropy & CKCE & JKCE & ECE \\ \hline
\multirow{3}{*}{ImageNet} & ResNet & 0.804 & 0.938 & 0.061 & 0.002 & 0.316 \\
 & ViT & 0.851 & 0.652 & 0.044 & 0.002 & 0.389 \\
 & Marginal & 0.001 & 6.908 & 0.000 & -0.002 & 0.003 \\
\bottomrule
    \end{tabular}
    \label{tab:nn_calibration}
\end{table*}

\subsection{Robustness Against Differences in Marginal Distribution of Predictions}\label{sec:experiments_1}

We begin by comparing the calibration error of models with different marginal distribution of predictions on the ImageNet dataset. We consider two pretrained neural network image classifiers, ResNet and ViT available from the \texttt{timm} library\footnote{The specific parameters used are \texttt{resnet50.a1\_in1k} and \texttt{ViT-L-16-SigLIP-384}}. In addition, we also compared the marginal probability model which predicts equal probability across all labels. Table \ref{tab:nn_calibration} shows the accuracy, cross entropy and calibration error (CKCE, JKCE and ECE) for the models computed on the validation data. Although all calibration error metrics indicate that the marginal distribution model has the lowest calibration error (JKCE produces a negative value as it is an unbiased estimator), there is a disagreement regarding which neural network is best calibrated. CKCE indicates that the ViT model has lower calibration error which agrees with the findings of \cite{Minderer_2021_revisiting} that transformer based models typically have better calibration than those with convolutional layers. On the other hand, JKCE ranks the models equally while ECE has ResNet as superior.

Figure \ref{fig:reliability-diag} provides more insight into why this occurs. The histograms show that most ResNet predictions lie in the corners of the probability simplex, whereas ViT is more conservative. Indeed the reliability diagrams in the same plot show that ResNet is overconfident, while ViT is slightly underconfident. Despite the reliability diagram for ViT appearing to have a closer match between model confidence and accuracy, particularly in the interval [0.5, 1], ResNet is preferred in terms of ECE as it makes more predictions in regions of better calibration. It is in some sense rewarded for its overconfidence. However, this may not be ideal for practitioners who would like for all outputs to be well calibrated. In contrast, our CKCE metric is unaffected by the marginal distribution of outputs, and thus provides a more reliable measure of calibration. 

\begin{figure*}[t]
    \centering
    \includegraphics[width=0.44\linewidth]{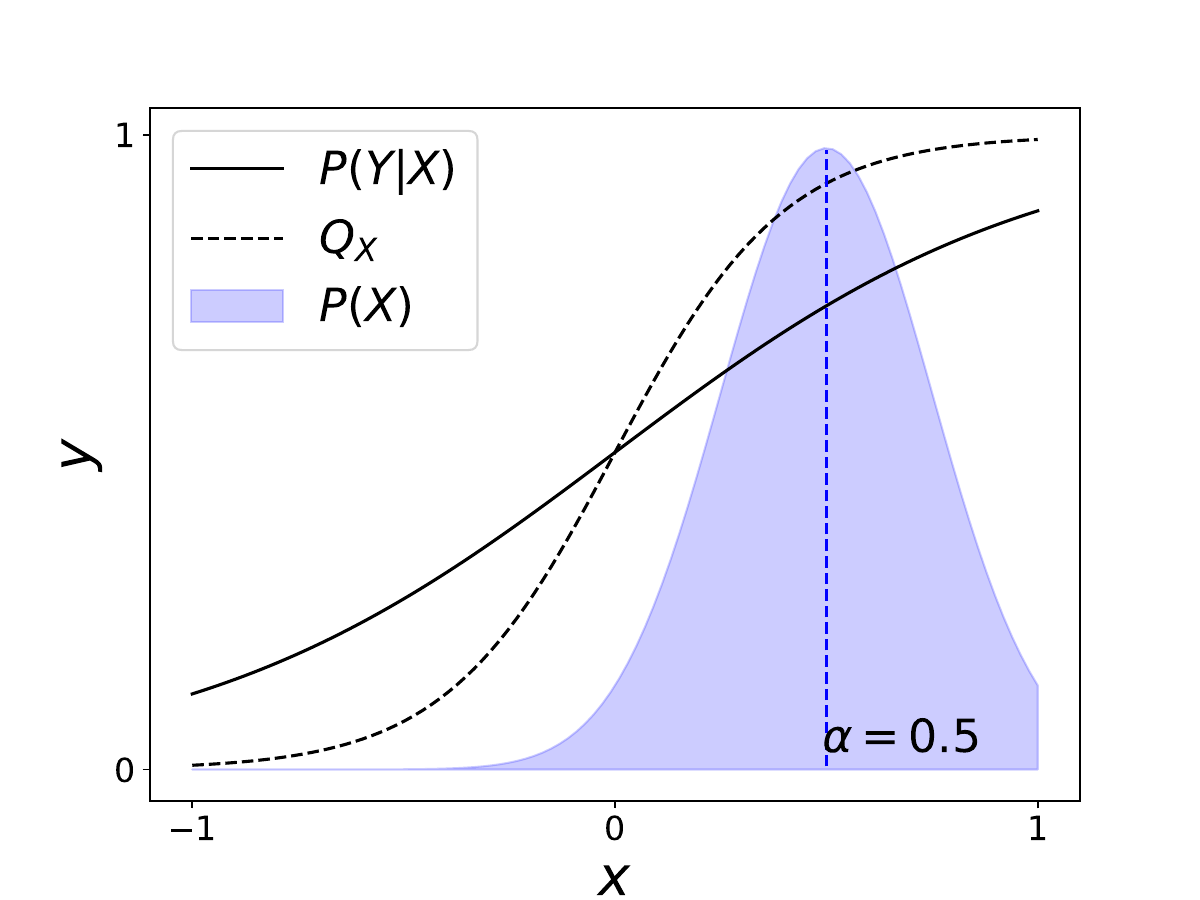}
    \includegraphics[width=0.44\linewidth]{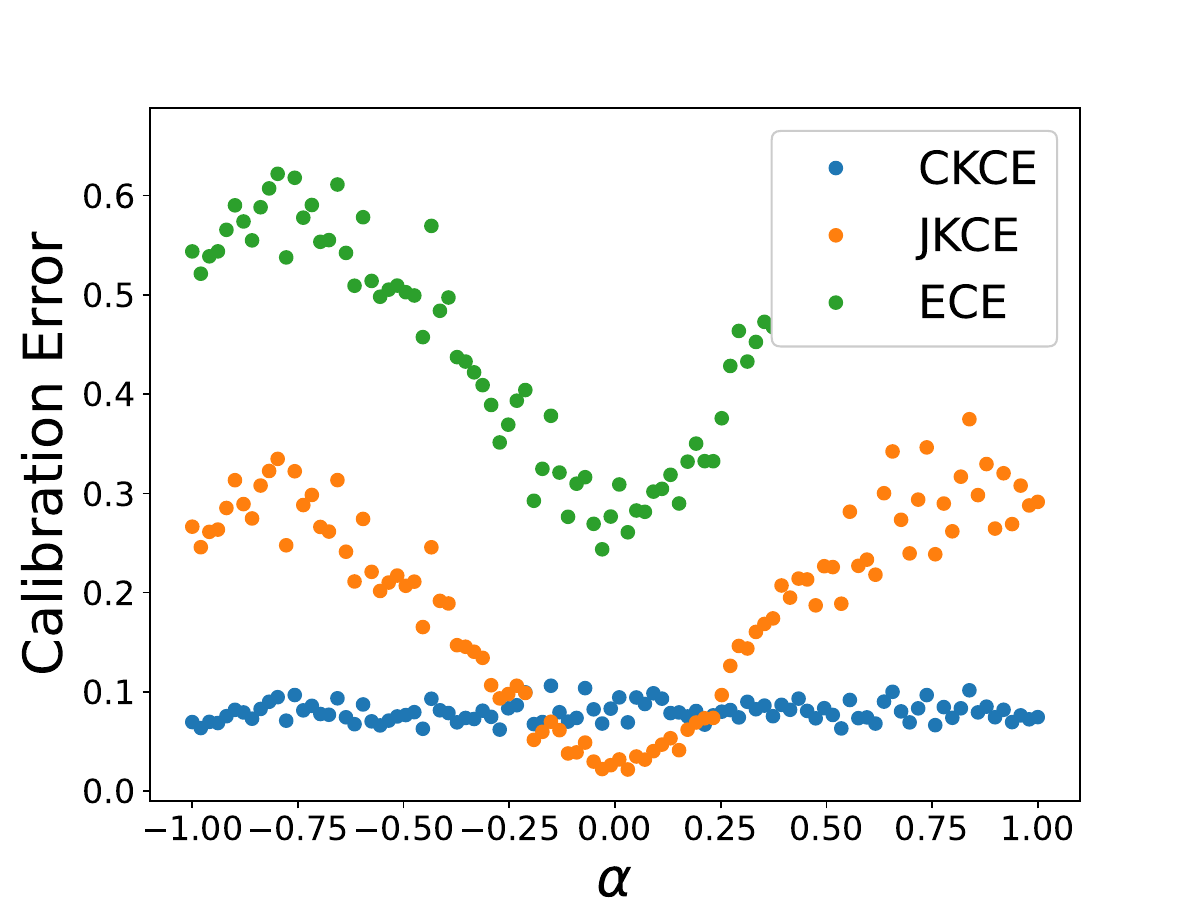}
    \caption{The CKCE estimator remains stable under covariate shift, whereas JKCE and ECE are highly sensitive to changes in the input distribution.}
    \label{fig:covshift}
\end{figure*}

\subsection{Robustness Against Covariate Shift}

\subsubsection{Synthetic Data} \label{sec:covshift_synthetic}

Let $\mathcal{X}\times \mathcal{Y} = [-1, 1] \times \{0,1\}$ with a conditional distribution governed by $P(Y=1|X=x) = \frac{1}{1+e^{-x}}$. Next, we choose a (miscalibrated) probabilistic model $f(x) = \frac{1}{1+e^{-5x}}$. We let the marginal distribution $P(X)$ be a truncated normal with domain $[-1,1]$, scale parameter $\sigma = 0.25$ and location parameter $\alpha$ which we vary between $-1$ and $1$. Note that the conditional distribution $P(Y|X)$ remains unchanged, making this an example of covariate shift. The set up is illustrated in Figure \ref{fig:covshift} (left). At each value of $\alpha$, we sample $n=1000$ inputs from $P(X)$ and simulate labels using $P(Y|X)$. The resulting calibration set is used to estimate each of CKCE, JKCE and ECE. 

Even though a change in $P(X)$ changes the marginal distribution of $Q_X$, the conditional distribution $P(Y|Q_X)$ and $P(Z_X|Q_X)$ are not effected. Therefore, we would like for that calibration error to stay constant as the location parameter $\alpha$ changes. This would indicate that the calibration error estimator is robust against covariate shift. The results in Figure \ref{fig:covshift} (right) show that the CKCE estimator indeed stays constant, with some scattering due to randomness in the generated samples, demonstrating the desired behaviour. On the other hand, both JKCE and ECE are highly sensitive to changes in $P(X)$, showing clear variability as the parameter $\alpha$ shifts.

\subsubsection{ImageNet} \label{sec:covshift_imagenet}

We next evaluate the impact of distribution shift on the calibration of image models, using the ResNet and ViT from Section \ref{sec:experiments_1}. Their performance on the full validation set is displayed in Table \ref{tab:nn_calibration}. Here we take a subset of $n=1000$ validation data, and alter the brightness of the input image, ranging from -50\% to +50\% in increments of 10\%. At each brightness level, the CKCE and JKCE for both models is computed. 
Due to the large number of classes, the ECE binning scheme suffers from the curse of dimensionality and cannot be accurately estimated.

\begin{figure}[t]
    \centering
    \includegraphics[width=0.9\linewidth]{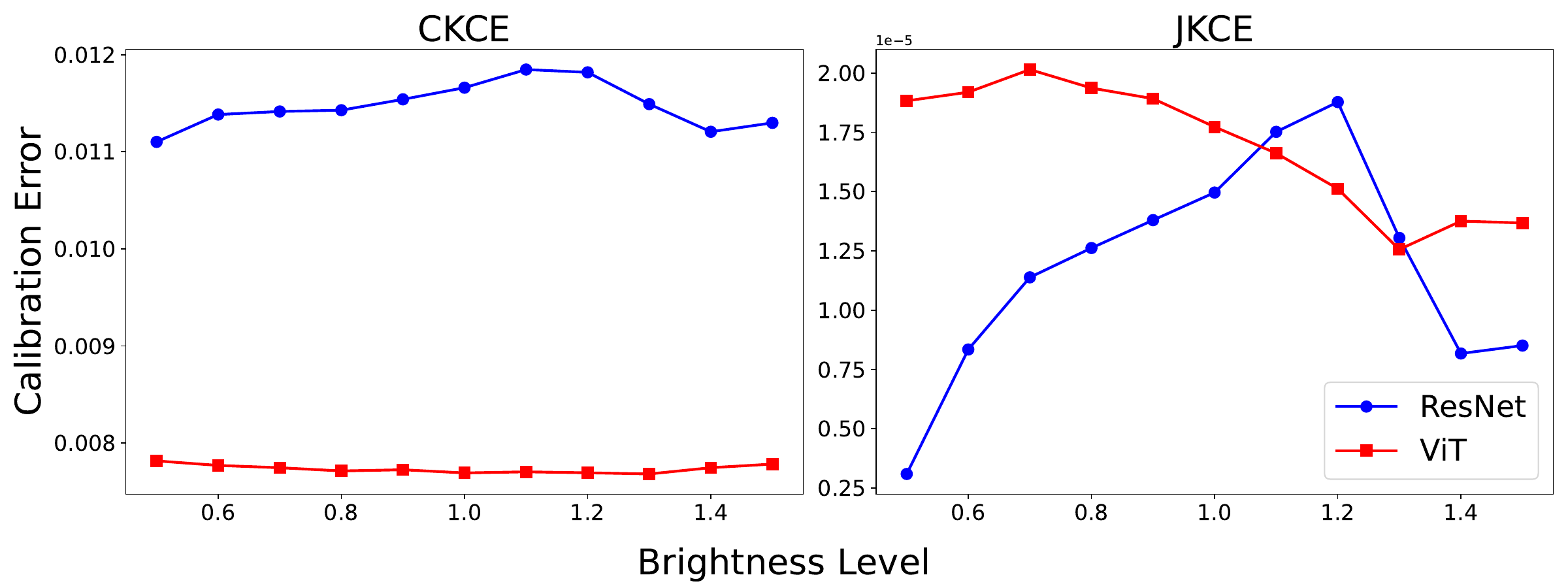}
    \caption{Calibration error of models with changing image brightness. CKCE provides consistent model preference, unlike JKCE.}
    \label{fig:imagenet_brightness}
\end{figure}

As the brightness changes, we would expect that model miscalibration, and hence relative ranking of the models, remains fairly constant. While an alteration in image brightness may cause a shift in the marginal distribution of $Q_X$, the conditional distribution $P(Y|Q_X)$ should be stable. Indeed, Figure \ref{fig:imagenet_brightness} shows that CKCE consistently indicates that ViT has lower calibration error than ResNet, with the estimate for both models largely unaffected by the change in input image brightness. On the other hand, JKCE is highly sensitive to the shift in input distribution, with the estimator preferring different models for different brightness levels. 

\subsection{Kernel choice robustness} \label{sec:robustness}

\begin{figure}[t]
    \centering
    \includegraphics[width=0.7\linewidth]{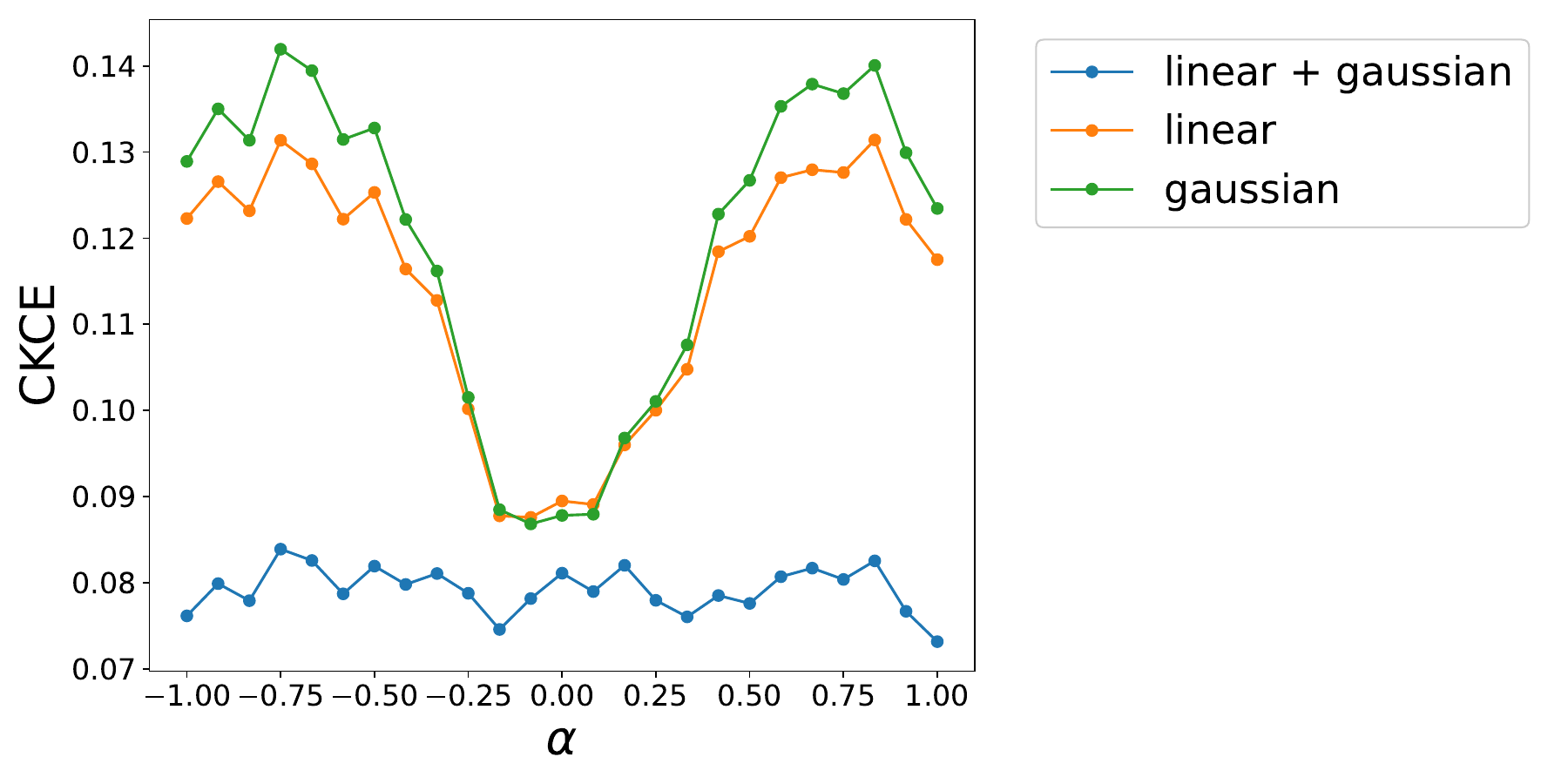}
    \caption{CKCE of a probabilistic model for three choices of kernel on the input variable. Using a kernel that combines a linear and a Gaussian component provides a more robust measure of calibration error.}
    \label{fig:kernel_robustness}
\end{figure}

We conclude with an experiment that demonstrates the necessity for the choice of kernel in Section \ref{sec:choice_kernel}. Using the same synthetic data setting as in Section \ref{sec:covshift_synthetic}, we compare the kernel (\ref{eq:kernel}) against simply using a linear or Gaussian kernel. As before, we range the parameter $\alpha$ between -1 and 1 at 25 discrete values and calculate the CKCE using the three choices of kernel $k$. The mean of 20 trials is plotted in Figure \ref{fig:kernel_robustness}. 

As we saw in the previous section, computing the CKCE with the combined linear-Gaussian kernel provides a metric that is robust against covariate shift. On the other hand, when a linear or Gaussian kernel are used individually on the input variable, then the CKCE no longer remains consistent, with a noticeable drop near $\alpha = 0$. The calibration error plot suggests that kernel (\ref{eq:kernel}) leads to CMO estimators that are less sensitive to the marginal distribution of inputs.

\section{Conclusion}

We introduced the CKCE - a novel method of measuring calibration based on conditional mean operators. We demonstrated that the CKCE provides a more reliable way of performing a relative comparison of calibration of probabilistic classifiers compared to other metrics. It is less effected by differences in marginal distributions of model outputs and remains stable under covariate shift. While we do not address ways to improve model calibration in this work, future research could explore using CKCE as a regularisation term in model training. Its ability to more precisely capture relative comparisons as training progresses may offer advantages over other metrics.

\begin{credits}
\subsubsection{\ackname} The authors would like to express their gratitude to Jiawei Liu and Russell Tsuchida for providing helpful guidance and discussion in the early stages of this work.

\subsubsection{\discintname}
The authors have no competing interests to declare that are relevant to the content of this article.
\end{credits}


\bibliographystyle{splncs04}
\bibliography{bibliography}

@inproceedings{ChaBouSej2021,
  title = {{Deconditional Downscaling with Gaussian Processes}},
  author = {Chau, Siu Lun and Bouabid, Shahine and Sejdinovic, Dino},
  year = {2021},
  volume = {34},
  pages = {17813--17825},
  booktitle = {Advances in Neural Information Processing Systems}
}

@article{phan2025hle,
title={Humanity's Last Exam},
author={{Phan et al.}},
journal={arXiv},
year={2025}
}

@article{liUnifiedAnalysisRandom2021,
  title = {Towards {{A Unified Analysis}} of {{Random Fourier Features}}},
  author = {Li, Zhu and Ton, Jean-Francois and Oglic, Dino and Sejdinovic, Dino},
  year = {2021},
  journal = {Journal of Machine Learning Research},
  volume = {22},
  number = {108},
  pages = {1--51}
}

@article{sriperumbudurUniversalityCharacteristicKernels2011,
  title = {Universality, {{Characteristic Kernels}} and {{RKHS Embedding}} of {{Measures}}},
  author = {Sriperumbudur, Bharath K. and Fukumizu, Kenji and Lanckriet, Gert R. G.},
  year = {2011},
  journal = {Journal of Machine Learning Research},
  volume = {12},
  number = {70},
  pages = {2389--2410}
}

@inproceedings{rahimiRandomFeaturesLargescale2007,
  title = {Random Features for Large-Scale Kernel Machines},
  booktitle = {Advances in {{Neural Information Processing Systems}} },
  author = {Rahimi, Ali and Recht, Benjamin},
  year = {2007},
  series = {20}
}

@article{muandetKernelMeanEmbedding2017a,
  title = {Kernel {{Mean Embedding}} of {{Distributions}}: {{A Review}} and {{Beyond}}},
  shorttitle = {Kernel {{Mean Embedding}} of {{Distributions}}},
  author = {Muandet, Krikamol and Fukumizu, Kenji and Sriperumbudur, Bharath and Sch{\"o}lkopf, Bernhard},
  year = {2017},
  month = jun,
  journal = {Foundations and Trends{\textregistered} in Machine Learning},
  volume = {10},
  number = {1-2},
  pages = {1--141}
}

@article{Brier_1950,
    author = {Brier, Glen W.},
    title = {Verification of forecasts expressed in terms of probability},
    journal = {Monthly Weather Review},
    volume = {78},
    number = {1},
    pages = {1--3},
    year = {1950}
}

@article{DeGroot_1983,
 author = {Morris H. DeGroot and Stephen E. Fienberg},
 journal = {Journal of the Royal Statistical Society. Series D (The Statistician)},
 number = {1/2},
 pages = {12--22},
 title = {The Comparison and Evaluation of Forecasters},
 volume = {32},
 year = {1983}
}

@InProceedings{Guo_2017,
  title = 	 {On Calibration of Modern Neural Networks},
  author =       {Chuan Guo and Geoff Pleiss and Yu Sun and Kilian Q. Weinberger},
  booktitle = 	 {Proceedings of the 34th International Conference on Machine Learning},
  year = 	 {2017}
}

@InProceedings{naeini_2015,
	author = {Naeini, Mahdi Pakdaman and Cooper, Gregory and Hauskrecht, Milos},
	booktitle = {Proceedings of the AAAI Conference on Artificial Intelligence},
	title = {Obtaining {Well} {Calibrated} {Probabilities} {Using} {Bayesian} {Binning}},
	series = {29},
	year = {2015}
}

@InProceedings{Nixon_2019,
    author = {Nixon, Jeremy and Dusenberry, Michael W. and Zhang, Linchuan and Jerfel, Ghassen and Tran, Dustin},
    title = {Measuring Calibration in Deep Learning},
    booktitle = {Proceedings of the IEEE/CVF Conference on Computer Vision and Pattern Recognition (CVPR) Workshops},
    month = {June},
    year = {2019}
}

@inproceedings{vaicenavicius_2019,
	author = {Vaicenavicius, Juozas and Widmann, David and Andersson, Carl and Lindsten, Fredrik and Roll, Jacob and Sch{\"o}n, Thomas},
	booktitle = {Proceedings of the 22nd {International} {Conference} on {Artificial} {Intelligence} and {Statistics}},
	title = {Evaluating model calibration in classification},
	year = {2019}
}

@inproceedings{Widmann_2019,
	author = {Widmann, David and Lindsten, Fredrik and Zachariah, Dave},
	booktitle = {Advances in Neural Information Processing Systems},
	title = {Calibration tests in multi-class classification: A unifying framework},
	series = {32},
	year = {2019}
}

@article{Gretton_2012,
  author  = {Arthur Gretton and Karsten M. Borgwardt and Malte J. Rasch and Bernhard Sch{{\"o}}lkopf and Alexander Smola},
  title   = {A Kernel Two-Sample Test},
  journal = {Journal of Machine Learning Research},
  year    = {2012},
  volume  = {13},
  number  = {25},
  pages   = {723-773}
}

@inproceedings{Song_2009,
author = {Song, Le and Huang, Jonathan and Smola, Alex and Fukumizu, Kenji},
title = {Hilbert space embeddings of conditional distributions with applications to dynamical systems},
year = {2009},
booktitle = {Proceedings of the 26th Annual International Conference on Machine Learning}
}

@InProceedings{Song_2010,
  title = 	 {Nonparametric Tree Graphical Models},
  author = 	 {Song, Le and Gretton, Arthur and Guestrin, Carlos},
  booktitle = 	 {Proceedings of the 13th International Conference on Artificial Intelligence and Statistics},
  year = 	 {2010}
}

@inproceedings{Park_2020,
	author = {Park, Junhyung and Muandet, Krikamol},
	booktitle = {Advances in Neural Information Processing Systems},
	title = {A Measure-Theoretic Approach to Kernel Conditional Mean Embeddings},
	series = {33},
	year = {2020}
}

@inproceedings{Ren_2016,
	author = {Ren, Yong and Zhu, Jun and Li, Jialian and Luo, Yucen},
	booktitle = {Advances in Neural Information Processing Systems},
	title = {Conditional Generative Moment-Matching Networks},
	series = {29},
	year = {2016}
}

@inproceedings{Li_2022,
	author = {Li, Zhu and Meunier, Dimitri and Mollenhauer, Mattes and Gretton, Arthur},
	booktitle = {Advances in Neural Information Processing Systems},
	title = {Optimal Rates for Regularized Conditional Mean Embedding Learning},
	series = {35},
	year = {2022}
}

@article{singh_2023,
	author = {Singh, R and Xu, L and Gretton, A},
	journal = {Biometrika},
	number = {2},
	pages = {497--516},
	title = {Kernel methods for causal functions: dose, heterogeneous and incremental response curves},
	volume = {111},
	year = {2023}
}

@inproceedings{Grunewalder_2012,
author = {Gr\"{u}new\"{a}lder, Steffen and Lever, Guy and Baldassarre, Luca and Patterson, Sam and Gretton, Arthur and Pontil, Massimilano},
title = {Conditional mean embeddings as regressors},
year = {2012},
booktitle = {Proceedings of the 29th International Conference on Machine Learning}
}

@article{Fine_2002,
author = {Fine, Shai and Scheinberg, Katya},
title = {Efficient {SVM} training using low-rank kernel representations},
year = {2002},
volume = {2},
number = {Dec},
journal = {Journal of Machine Learning Research},
pages = {243–264}
}

@inproceedings{Widmann_2021,
    title={Calibration tests beyond classification},
    author={David Widmann and Fredrik Lindsten and Dave Zachariah},
    booktitle={Proceedings of the 9th International Conference on Learning Representations},
    year={2021}
}

@article{Lee_2023,
  author  = {Donghwan Lee and Xinmeng Huang and Hamed Hassani and Edgar Dobriban},
  title   = {T-Cal: An Optimal Test for the Calibration of Predictive Models},
  journal = {Journal of Machine Learning Research},
  year    = {2023},
  volume  = {24},
  number  = {335},
  pages   = {1--72}
}

@InProceedings{Glaser_2023,
  title = 	 {Fast and scalable score-based kernel calibration tests},
  author =       {Glaser, Pierre and Widmann, David and Lindsten, Fredrik and Gretton, Arthur},
  booktitle = 	 {Proceedings of the 39th Conference on Uncertainty in Artificial Intelligence},
  year = 	 {2023}
}

@misc{Chatterjee_2024,
	author = {Chatterjee, Anirban and Niu, Ziang and Bhattacharya, Bhaswar B.},
	note = {arXiv 2407.16550},
	title = {A {Kernel}-{Based} {Conditional} {Two}-{Sample} {Test} {Using} {Nearest} {Neighbors} (with {Applications} to {Calibration}, {Regression} {Curves}, and {Simulation}-{Based} {Inference})},
	year = {2024}
}

@article{Fukumizu_2004,
	author = {Fukumizu, Kenji and Bach, Francis R. and Jordan, Michael I.},
	journal = {Journal of Machine Learning Research},
	number = {Jan},
	pages = {73--99},
	title = {Dimensionality {Reduction} for {Supervised} {Learning} with {Reproducing} {Kernel} {Hilbert} {Spaces}},
	volume = {5},
	year = {2004}
}

@inproceedings{Fukumizu_2007,
	author = {Fukumizu, Kenji and Gretton, Arthur and Sun, Xiaohai and Sch\"{o}lkopf, Bernhard},
	booktitle = {Advances in Neural Information Processing Systems},
	title = {Kernel Measures of Conditional Dependence},
	series = {20},
	year = {2007}
}

@article{Szabo_2018,
  author  = {Zolt{{\'a}}n Szab{{\'o}} and Bharath K. Sriperumbudur},
  title   = {Characteristic and Universal Tensor Product Kernels},
  journal = {Journal of Machine Learning Research},
  year    = {2018},
  volume  = {18},
  number  = {233},
  pages   = {1--29}
}

@article{Merkle_2013,
author = {Merkle, Edgar and Steyvers, Mark},
year = {2013},
month = {12},
pages = {292-304},
title = {Choosing a Strictly Proper Scoring Rule},
volume = {10},
journal = {Decision Analysis}
}

@inproceedings{Minderer_2021_revisiting,
title={Revisiting the Calibration of Modern Neural Networks},
author={Matthias Minderer and Josip Djolonga and Rob Romijnders and Frances Ann Hubis and Xiaohua Zhai and Neil Houlsby and Dustin Tran and Mario Lucic},
booktitle={Advances in Neural Information Processing Systems},
year={2021}
}

\end{document}